\documentclass{article}
\usepackage{spconf,amsmath,graphicx}
\usepackage[utf8]{inputenc}
\usepackage{bbm}
\usepackage{amssymb}
\usepackage{caption}
\usepackage{graphics}
\usepackage{subcaption}
\usepackage{csquotes}
\usepackage{amsthm, amsfonts}
\usepackage{breqn}
\usepackage{array,multirow}
\usepackage{algorithm}
\usepackage{algorithmic}
\usepackage{booktabs}
\newtheorem{theorem}{Theorem}
\usepackage{tablefootnote}
\usepackage{soul}
\usepackage{xcolor}
\usepackage{spconf}
\title{XMOL: Explainable Multi-property Optimization of Molecules}

\name{Aye Phyu Phyu Aung$^{1\star}$ 
\qquad Jay Chaudhary$^{\star \dagger}$ \qquad Ji Wei Yoon$^{\star}$ \qquad Senthilnath Jayavelu$^{1\star}$ 
\thanks{$^{1}$ corresponding authors.}
 \thanks{A.P.P.A., J.W.Y., and S.J. acknowledge funding from the Accelerated Materials Development for Manufacturing Program at A*STAR via the AME Programmatic Fund by the Agency for Science, Technology and Research under Grant No. A1898b0043 and J.C. acknowledge Singapore International Pre-Graduate Award (SIPGA) scholarship, A*STAR.}
}
\vspace{-0.8cm}
\address{$^{\star}$ Institute for Infocomm Research (I$^{2}$R), Agency for Science, Technology and Research (A*STAR), \\ 1 Fusionopolis Way, \#21-01, Connexis South Tower, Singapore, 138622 \\
$^{\dagger}$Department of Electrical Engineering, Indian Institute of Technology Bombay, India, 400076
\\
}
\begin{document}
\maketitle
\begin{abstract}
Molecular optimization is a key challenge in drug discovery and material science domain, involving the design of molecules with desired properties. Existing methods focus predominantly on single-property optimization, necessitating repetitive runs to target multiple properties, which is inefficient and computationally expensive. Moreover, these methods often lack transparency, making it difficult for researchers to understand and control the optimization process. To address these issues, we propose a novel framework, E\textbf{X}plainable \textbf{M}ulti-property \textbf{O}ptimization of Mo\textbf{L}ecules (XMOL), to optimize multiple molecular properties simultaneously while incorporating explainability. Our approach builds on state-of-the-art geometric diffusion models, extending them to multi-property optimization through the introduction of spectral normalization and enhanced molecular constraints for stabilized training. Additionally, we integrate interpretive and explainable techniques throughout the optimization process. We evaluated XMOL on the real-world molecular datasets i.e., QM9, demonstrating its effectiveness in both single property and multiple properties optimization while offering interpretable results, paving the way for more efficient and reliable molecular design.
\end{abstract}
\section{Introduction}
Molecular optimization is a pivotal task in drug discovery and material science domain, involving the design of molecules with desirable properties. It can be broadly classified into two categories: unconditional optimization and conditional optimization. Unconditional optimization focuses on generating molecules without any specific constraints, aiming for overall improved characteristics. In contrast, conditional optimization targets molecules that fulfill predefined criteria for particular physical, chemical, or biological properties within each molecular dataset.

State-of-the-art works in molecular optimization have employed various advanced techniques. JT-VAE~\cite{jin2018junction} represents molecules as junction trees to capture hierarchical structures; Mol-CycleGAN~\cite{maziarka2020mol} uses cycle-consistent adversarial networks to optimize molecular properties; and GCDM~\cite{morehead2023geometry} focuses on generating 3D molecular structures while preserving geometric constraints. Other significant contributions include VAEs for encoding molecular structures into continuous latent spaces, Grammar VAE~\cite{kusner2017grammar} for ensuring syntactically valid molecular structures, and MolGAN~\cite{de2018molgan} that combine GANs and reinforcement learning for property optimization. All these existing approaches primarily focus on optimizing one property at a time. Each run consumes significant computational resources, and the iterative process does not guarantee the simultaneous improvement of all desired properties. Therefore, a more efficient approach is needed to optimize multiple properties concurrently. Moreover, the complexity of molecular optimization models often leads to a lack of transparency in the optimization process. EXplainable AI (XAI) can provide researchers insidght into more interpretability and transparent optimization process, verify the results, and fine-tune the models effectively. 

To address these inefficiencies and the need for transparency, we propose a novel approach named E\textbf{X}plainable \textbf{M}ulti-property \textbf{O}ptimization of Mo\textbf{L}ecules (XMOL). Our contributions are three fold:

\noindent 1. Multi-property optimization by extending the SOTA single property geometric complete diffusion methods.

\noindent 2. We propose spectral normalization and new molecular constraints to stabilize the training process.

\noindent 3. An integrated architecture for multi-property molecular optimization with ante-hoc explainable design for transparent training process.

We evaluated our proposed method using real molecular datasets such as QM9, demonstrating its effectiveness in optimizing multiple properties concurrently while providing interpretable results.

\section{Related Works}

In this section, we will discuss the existing works for theory references and discussions on their shortcomings.

\subsection{Generative Models for Molecular Optimization}
Since geometric deep learning (GDL) ~\cite{cao2020comprehensive} is established with theoretical advances, researchers propose many novel applications in different fields. Moreover, generative modelling has also seen a significant advancement in the space of 2D/3D image generation~\cite{hong2023lrm,blattmann2022retrieval, rombach2022high, ho2020denoising, nichol2021improved}, computational biology, chemistry, drug discovery and physics~\cite{watson2023novo,xu2022geodiff,senthilnath2024self, mudur2022can}. Combining the two disciplines, molecular optimization is performed using two major methods: reinforcement learning and deep learning. RL approaches like the GCPN~\cite{you2018graph} optimize molecular properties by learning from environment interactions. GTRL \cite{polykovskiy2022generative} proposes automated drug design capabilities through deep learning, enabling rapid identification of potent molecules. However, they are complex and data-intensive. On the other hand, we saw advancements in all types of deep learning models such as VAEs, GANs and diffusion models. JT-VAE \cite{jin2018junction} represents molecules as junction trees to capture hierarchical structures. Mol-CycleGAN \cite{maziarka2020mol} and MolGAN \cite{de2018molgan} use cycle-consistent adversarial networks and RL-assisted GAN network to optimize molecular properties. Lastly, GCDM~\cite{morehead2023geometry} ensures geometric constraints in 3D molecule generation, addressing structural generation. However, all these approaches optimize one property at a time, inefficient for multi-property requirements making them resource intensive and not scalable. Hence, we extend the latest generative model to propose a multi-property optimization by doing necessary modifications to both property discriminator EGNN~\cite{satorras2021n} and generative model GCDM to be able to be competitively trained for multi-property molecular optimization as compared to single property optimization runs.

\subsection{Explainable Design for Molecular Structures}
To provide more insight to single property optimization process, S-REINFORCE~\cite{dutta2024interpretable} generates symbolic policies. Similarly, we introduce explainable AI (XAI) to the multi-property optimization. XAI can be categorized into 1) ante-hoc which integrate interpretability into the model design itself and 2) post-hoc methods which are for complex, often opaque models to provide explanations for their predictions after training. Ante-hoc methods~\cite{garg2024advancing,nagisetty2020xai} are usually decision-tree, rule-based and linear methods for efficiency. This simplicity has a trade-off and they struggle to explain graphical structures. Meanwhile, post-hoc methods such as LIME \cite{ribeiro2016should}, SHAP \cite{lundberg2017unified} and Saliency Maps \cite{simonyan2013deep} models use approximation, coorperative game theory and maps respectively to explain the contribution of each feature in prediction. However, they do not directly translate to graph-based data where the importance of a node or edge may depend on its neighbors and the overall graph topology. Hence, we extend the GNN classification explainable design, GNNSHAP \cite{akkas2024gnnshap}, modify it to be able to work as a regression explainable design and integrate it to our proposed architecture. 

\section{XMOL}

XMOL architecture mainly consists of 3 modules: multi-property geometric-complete diffusion model, multi-property EGNN and GNNSHAP for explainability. Our training objective is to optimize the combination of 3 losses:
\small
\begin{equation}
    \mathcal{L} = \mathcal{L}_{DM} + \mathcal{L}_{EGNN}+ \mathcal{L}_{Fidelity},
\end{equation}
\normalsize
where $\mathcal{L}_{DM}, \mathcal{L}_{EGNN}$ and $\mathcal{L}_{Fidelity}$ stand for the loss functions from each different module of the XMOL model, which will be defined in the respective subsections below.

\subsection{Constrained Conditional Optimization}
\noindent \textbf{Conditional Optimization.}
The first term $\mathcal{L}_{DM}$ represents the reconstruction loss of geometric-complete diffusion model~\cite{morehead2023geometry}, which uses the geometric-complete GCPNET++ as the denoising network of the molecular diffusion model. With $\hat{\epsilon}_t$ as the predicted noise from the denoising network, the optimization objective is defined as:
\small
\begin{equation}
    \mathcal{L}_{DM} =  \mathbb{E}_{\epsilon_t \sim \mathcal{N}_{x_h}(0,1)} \left[ \frac{1}{2} \| \epsilon_t - \hat{\epsilon}_t \|^2 \right],
\end{equation}
\normalsize

\noindent \textbf{Adding Constraint.} To improve the integrity and validity of the generated molecules, we add a Quantitative Estimate of Druglikeness (QED) as the regularization term. We use the desirability functions $d_{i}$ of the properties such as Molecular Weight (MW),  Octanol-water partition coefficient (LogP), Number of hydrogen bond donors (HBD), Number of hydrogen bond acceptors (HBA), Molecular polar surface area (PSA), Number of rotatable bonds (ROTB), aromatic rings (AROM) and structural alerts (ALERTS) as defined by \cite{bickerton2012quantifying} to calculate QED value as defined by:
\vspace*{-0.2cm}
\small
\begin{equation}
    \text{QED} = \exp\left(\frac{1}{n} \sum_{i=1}^{n} \ln d_i\right),
\end{equation}
\normalsize
where $n$ is the total number of properties used to calculate QED. We add the QED loss to $\mathcal{L}_{DM}$ by setting the threshold $\tau= 0.5$ as follows:
\vspace*{-0.2cm}
\small
\begin{equation}
    \text{QED\_loss} = \begin{cases}
        0  & \text{, if } \text{QED} > \tau \\
        (\text{QED} - \tau)^{2} &, \text{otherwise}.
    \end{cases}, 
\end{equation}
\normalsize

\subsection{Multi-property predicting EGNN}
The second term $\mathcal{L}_{EGNN}$ is the regression loss for each property of the generated molecule. Referencing EGNN~\cite{satorras2021n}, we represent the molecule with a graph structure $G= (V,E)$ where $v_{i} \in V$ represent the atoms and $e_{ij} \in E$ represent the bond between them. We also define feature node embeddings $\mathbf{h}_{i} \in \mathtt{R}^{nf}$, n-dimensional coordinate embeddings $\mathbf{x}_{i} \in \mathtt{R}^{n}$ and feature edge embeddings $\mathbf{e}_{ij} \in \mathbf{R}^{ef}$. To adapt EGNN for predicting multi-property values, we modify the output layer to produce a vector of predicted properties adding $\hat{y}_{i}$. With $\phi_{e}, \phi_{x}, \phi_{h}$ as learnable functions and $\psi$ as non-linear activation function, the output layer is updated as:
\vspace*{-0.2cm}
\small
\begin{align}
    \mathbf{m}_{ij} &= \phi_{e} (\mathbf{h}_{i}, \mathbf{h}_{j}, \|\mathbf{x}^{l}_{i} - \mathbf{x}^{l}_{j} \|^{2}, \mathbf{e}_{ij}), 
    \\
    \mathbf{x}^{l+1}_{i} &= \mathbf{x}^{l}_{i} + C \sum_{j \neq i} ( \mathbf{x}^{l}_{i}- \mathbf{x}^{l}_{j}) \phi_{x} (m_{ij}),
    \\
    \mathbf{h}^{l+1}_{i} &= \phi_{h} (\mathbf{h}^{l}_{i}, \sum_{j \neq i} m_{ij}),
    \\ 
    \hat{y}_{i} &= \psi (\mathbf{h}^{L}_{i}),
\end{align}
\normalsize

where $l \in \mathcal{L}$ represent each layer of EGNN with $L$ being the last layer. The loss function is defined as the weighted sum of the mean absolute error (MAE) losses for each property:
\vspace*{-0.2cm}
\small
\begin{equation}
    \mathcal{L}_{EGNN} = \sum_{i=1}^{P} w_{i} MAE(y_{i}, \hat{y}_{i}).
\end{equation}
\normalsize

\noindent \textbf{Spectral Normalization.} 
To enhance the stability and performance of our EGNN for multi-property regression, we incorporate the spectral normalization~\cite{miyato2018spectral} as a regularization technique. The regularization term controls the Lipschitz constant of the model by normalizing the spectral norm of each layer's weight matrix to ensure that the model's activations do not explode or vanish. 

\begin{theorem}
Let \( f: \mathbb{R}^n \to \mathbb{R} \) be a function defined by a composition of layers in EGNN, such that \( f = g_k \circ \alpha_k \circ g_{k-1} \circ \dots \circ g_1 \), where each layer \( g_i \) represents a linear and $\alpha$ represents a non-linear function. Then, the spectral normalization process makes sure that:
\vspace*{-0.2cm}
\small
\begin{equation}
    \| f(x) \|_{\text{Lip}} \leq 1 \quad , \forall x \in \mathbb{R}^n.
\end{equation}
\normalsize
\end{theorem}
\noindent \textbf{Definition:} \textit{A function f is called K-Lipschitz continuous if $\lVert f(x)-f(x')\rVert/\lVert x-x'\rVert \leq K$ for any $x,x'$; where norm being the $l2$ norm.}
\vspace*{-0.2cm}
\begin{proof}
Spectral normalization~\cite{miyato2018spectral} regulates the Lipschitz constant of a layer $ g $ by constraining its spectral norm, where $ g: h_{in} \rightarrow h_{out} $. The Lipschitz norm $ \lVert g \rVert_{\text{Lip}} = \sigma(\sup_h(\Delta g(h))) $ can be calculated as:
\vspace*{-0.2cm}
\small
\begin{equation}
    \sigma(A) := \max\limits_{h,h\neq0}\frac{\lVert Ah\rVert_2}{\lVert h\rVert_2},
\end{equation}
\normalsize
\noindent where $\sigma(A)$ is the spectral norm of the matrix $ A $ i.e., the largest singular value of $A$. Therefore, the norm for a linear layer $g(h) = Wh$ is given by:
\vspace*{-0.2cm}
\small
\begin{equation}
\|g\|_{\text{Lip}} = \sup_h \sigma(\nabla g(h)) = \sup_h \sigma(W) = \sigma(W).
\label{normforlinearfunction}
\end{equation}
\normalsize
Since Lipschitz norm of the activation function $ \|\alpha\|_{\text{Lip}} $ is 1 (for SiLU, it is slightly greater than 1~\cite{nwankpa2018activation}), we can represent the norm of linear functions as:
\vspace*{-0.2cm}
\small
\begin{equation}
    \|g_1 \circ g_2\|_{\text{Lip}} \leq \|g_1\|_{\text{Lip}} \cdot \|g_2\|_{\text{Lip}},  
\end{equation}
\normalsize
We apply the spectral normalization to all linear layers within the node decoder and graph decoder, which are composed of linear layers and non-linear activation functions. \\
Let $ f : \mathbb{R}^n \to \mathbb{R}^m $ be a function representing these layers, defined by the composition 
$ f = g_k \circ \alpha_k \circ g_{k-1} \circ \cdots \circ g_1 $, where each layer $ g_i $ represents a linear function and 
$ \alpha $ represents a non-linear function.\\
We use SiLU activation function which has a Lipschitz constant slightly greater than 1 to observe the bound on $\|f\|_{\text{Lip}}$:
\vspace*{-0.2cm}
\small
\begin{align}
\|f\|_{\text{Lip}} &\leq \prod_{i=1}^{k} \|h_i \mapsto W_i h_i\|_{\text{Lip}} \cdot \prod_{i=1}^{k} \|\alpha_i\|_{\text{Lip}},
\\
&\approx \prod_{i=1}^{k} \|h_i \mapsto W_i h_i\|_{\text{Lip}},
\\
&\approx \prod_{i=1}^{k} \sigma(W_i).
\end{align}
\normalsize
Spectral normalization process scales the spectral norm of the weight matrix $W_i$ so that it satisfies the Lipschitz constraint $\sigma(W_i) = 1$:
\vspace*{-0.2cm}
\small
\begin{equation}
\bar{W}_{\text{SN}}(W_i) := \frac{W_i}{\sigma(W_i)}.
\end{equation}
\normalsize
Using the above normalization for each $ W_i $, we can appeal $\sigma\left(\bar{W}_{\text{SN}}(W_i)\right) = 1$. Hence, $\|f\|_{\text{Lip}}$ is upper-bounded by 1.
\vspace*{-0.2cm}
\small
\begin{equation}
    \|f\|_{\text{Lip}} \leq 1.
\end{equation}
\normalsize
\end{proof}

\subsection{Model Explainability for Property Regression}
For the reliability of the property optimization process by our proposed model, the third term $\mathcal{L}_{Fidelity}$ represents the fidelity loss to measure how accurate the model can reconstruct the architecture preserving all the property values that are predicted. For the explainability model, we adapt GNNShap~\cite{akkas2024gnnshap} to compute the Shapley values for features of the molecular structure where the edges are considered as players. We follow the steps of GNNShap to compute the Shapley values for each property prediction: 1) pruning the redundant edges; 2) sampling the coalitions (subgraphs) creating mask matrix $M$ and weight matrix $W$; 3) computing the EGNN prediction where $\hat{y}_{i}, i \in P$ and finally; 4) computing Shapley values for all edges in the computation graph by: 
\vspace*{-0.2cm}
\small
\begin{equation}
    \phi_{i} = (M^{T}WM)^{-1}M^{T}W\hat{y}_{i}.
\end{equation}
\normalsize

With $\hat{y}_{i}, i \in P$ from EGNN and $\phi_{i}$ from GNNShap, we compute  $\mathcal{L}_{Fidelity}$ by evaluating the model on a 
perturbed graph $G'$ where the least important edges of the molecular structure are removed. Hence, we obtain a new prediction $\hat{y}'_{i}$. We then calculate the fidelity loss by taking the difference in the prediction accuracy $A(.)$ between the original and perturbed models, as: 
\vspace*{-0.2cm}
\small
\begin{equation}
    \mathcal{L}_{Fidelity} = \sum_{i\in P} \left (A(\hat{y}_{i}) - A(\hat{y}'_{i}) \right).
\end{equation}
\normalsize

\begin{figure}[ht]
\centering
\begin{subfigure}[b]{0.22\textwidth}
\centering
\includegraphics[width=\textwidth]{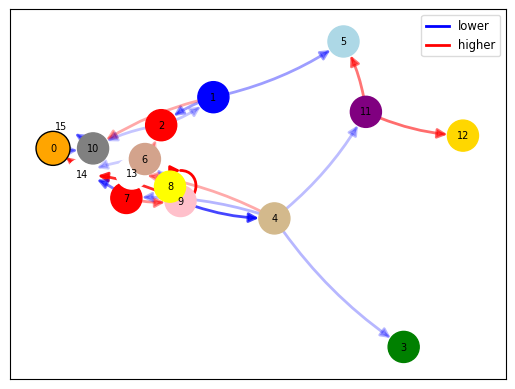}
\end{subfigure}
\hspace{1mm}
\begin{subfigure}[b]{0.22\textwidth}
\centering
\includegraphics[width=\textwidth]{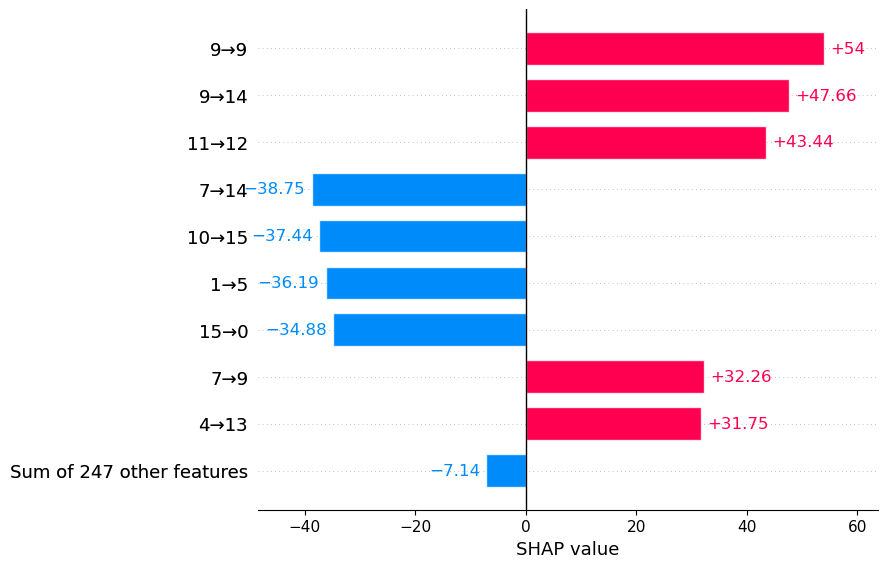}
\end{subfigure}
    \caption{Graph and bar plots illustrating the contributions by each edge of the generated molecule to EGNN prediction}
    \label{fig:graphbar}
\end{figure}

\begin{table*}[ht!]
\centering
\caption{Multi-property optimization and time efficiency for the ablations of XMOL}
\begin{tabular}{|l|c|c|c|c|c|c|c|}
\hline
\multirow{2}{*}{\textbf{Method}} & \textbf{$\alpha$ $\downarrow$} & \textbf{Gap $\downarrow$} & \textbf{HOMO $\downarrow$} & \textbf{LUMO $\downarrow$} & \textbf{$\mu$ $\downarrow$} & \textbf{$C_v$ $\downarrow$} & \textbf{Time $\downarrow$} \\
 & (Bohr$^3$) & (meV) & (meV) & (meV) & (D) & (cal.K/mol) & (hrs) \\
\hline
XMOL-S & 1.88 & 610 & 340 & 482 & 0.845 & 0.683& 559.73\\
Ours & 3.39 & 869 & 441 & 908 & 1.009 & 1.397 & 79.868 \\
Ours+QED & 3.1 & 901 & 457 & \underline{781} & \underline{1.05} & 1.346 & 145.716*\\
XMOL & \underline{3.04} & \underline{858} & \underline{432} & 817 & 1.078 & \underline{1.204} & \underline{92.963} \\
\hline
\end{tabular}

\footnotesize{* Run on Ubuntu 22.04 System with AMD Ryzen 9 5950X 16-Core CPU, 128GB RAM and Nvidia RTX A6000.}
\label{tab:timecomparison}
\end{table*}

\section{Experiments}
\noindent \textbf{Dataset.}  For our experiments, we use QM9 dataset~\cite{Ramakrishnan2014}. The dataset contains 134,000 stable small organic molecules and each molecule includes isotropic polarizability $\alpha$, HOMO, LUMO, gap, dipole moment $\mu$ and heat capacity $C_v$ as the properties we choose to optimize with XMOL.

\noindent \textbf{Baselines.}
In our experiments, we evaluated the performance of our proposed method against several baselines to demonstrate its efficacy and robustness in molecular optimization. The baselines include: GCDM~\cite{morehead2023geometry}, EDM~\cite{hoogeboom2022equivariant} and GeoLDM~\cite{xu2023geometric}. We first evaluated the single property version of our model (XMOL-S) for the fair comparison against the baselines. Then, we implemented the following ablations of our method (multi-property molecular optimization with spectral normalization) using: i) both QED constraint and fidelity loss (XMOL); ii) QED constraint only (Ours+QED); and iii) without both (Ours) to demonstrate the time efficiency of our method. We trained every baseline and ablation for 1500 epochs and for GCDM and our methods we use the QM9 validation split for training the conditional models and QM9 training split as the EGNN training dataset~\cite{hoogeboom2022equivariant}. For all the baselines, we report mean absolute error (MAE) of molecular property prediction. All our experiments unless specified are performed on Ubuntu 22.04 System with Intel i9-14900KF CPU, 64GB RAM and Nvidia GeForce RTX4080 Super.

\noindent \textbf{Visualizing with GNNShap.} Apart from calculating the fidelity loss, we also use GNNShap to illustrate the contribution by each node and edge feature to the EGNN regression with graph and bar plots. Figure~\ref{fig:graphbar} shows the graph and bar plots for a generated molecule with 16 atoms having property values $[\alpha = 0.4309, \textit{gap}= -2.1000,  \textit{homo}=2.4312, \textit{lumo}=-1.1288,  \mu =0.1149,  C_v= 1.1695]$.

\begin{table}[ht]
\centering
\scriptsize
\caption{Comparison of single property configuration of XMOL with different baselines.}
\begin{tabular}{|l|c|c|c|c|c|c|}
\hline
\multirow{2}{*}{\textbf{Method}} & 
\textbf{$\alpha$ $\downarrow$} & \textbf{Gap $\downarrow$} & \tiny\textbf{HOMO $\downarrow$} & \tiny\textbf{LUMO $\downarrow$} & \textbf{$\mu$ $\downarrow$} & \textbf{$C_v$ $\downarrow$}  \\ 
\tiny
 & \tiny(Bohr$^3$) & \tiny(meV) & \tiny(meV) & \tiny(meV) & \tiny(D) & \tiny(cal.K/mol) \\
\hline
 \scriptsize
GCDM~\cite{morehead2023geometry} & 1.97 & 602 & 344 & 479 & 0.844 & 0.689 \\
EDM~\cite{hoogeboom2022equivariant} & 2.76 & 655 & 356 & 584 & 1.111 & 1.101 \\
GeoLDM~\cite{xu2023geometric} & 2.37 & 587 & 340 & 522 & 1.108 & 1.025 \\
XMOL-S & \textbf{1.88} & \underline{610} & \textbf{340} & \underline{482} & \textbf{0.845}& 
\textbf{0.683}  \\
\hline
\end{tabular}
\footnotesize{The baselines results are as reported in ~\cite{morehead2023geometry}.}
\label{tab:comparison}
\end{table}

\noindent \textbf{Quantitative Evaluation.} Table~\ref{tab:comparison} compares the single-property performance of XMOL-S with baseline methods such as GCDM, EDM, and GeoLDM. XMOL-S excels with the lowest $\alpha$, HOMO, $C_v$ and $\mu$. Although it trails slightly in Gap and LUMO, the results remain competitive, with only minor differences from the baselines. Overall, XMOL-S demonstrates strong performance in optimizing all the properties, standing out against the well-established baselines.

\vspace{1mm}
\noindent \textbf{Computational Efficiency for Multi-property Optimization.} Table~\ref{tab:timecomparison} highlights the time efficiency of our method in optimizing multiple properties. XMOL-S takes 559.73 hrs for training, while the XMOL ablations significantly improve this: 79.868 hrs for the unconstrained version (Ours), 145.716 hrs* for the constrained version (Ours+QED), and just 92.963 hrs for the full XMOL. These results show that XMOL effectively balances time and performance while maintaining valid molecules. Therefore, we can  use XMOL as the primary method for training conditional models, with brief fine-tuning using XMOL-S for a few epochs to boost efficiency without sacrificing performance.

\vspace*{-0.2cm}
\section{Conclusion}
In this paper, we proposed XMOL, a novel framework for explainable multi-property optimization of molecules. XMOL extends SOTA geometry-complete diffusion models by incorporating spectral normalization and adding constraints to stabilize training for multi-property optimization. Additionally, we integrated explainable designs to provide interpretability and transparency. Our experiments demonstrate that XMOL is effective in optimizing multiple properties and offering reliable explanations. XMOL thus represents a significant advancement towards more efficient and transparent conditional molecular optimization. Future work could explore extending XMOL to handle larger and more complex molecular structures, as well as integrating additional property-specific constraints to further improve optimization outcomes.

\clearpage
\bibliographystyle{IEEEbib}
\bibliography{refs}


\end{document}